%% file: anonymous-submission-latex-2026.tex
\documentclass[letterpaper]{article} 
\usepackage{aaai2026}  
\usepackage{times}  
\usepackage{helvet}  
\usepackage{courier}  
\usepackage[hyphens]{url}  
\usepackage{graphicx} 
\usepackage{natbib}  
\usepackage{caption} 
\usepackage{algorithm}
\usepackage{algorithmic}

\usepackage{algorithm}
\usepackage{algorithmic}

\usepackage{subfigure}
\usepackage{amsmath}
\usepackage{amsfonts}
\usepackage{booktabs}
\usepackage{enumitem}

\usepackage{graphicx} 
\usepackage{multirow} 
\usepackage{booktabs}
\usepackage{amsmath}
\usepackage{makecell}
\usepackage{caption}
\usepackage{amssymb}

\usepackage{subfigure}
\usepackage{graphicx}
\usepackage{algorithm}
\usepackage{algorithmic}
\usepackage{bbding}
\usepackage{booktabs}
\usepackage{tcolorbox}

\usepackage{amssymb}

\usepackage{newfloat}
\usepackage{listings}
\DeclareCaptionStyle{ruled}{labelfont=normalfont,labelsep=colon,strut=off} 
\floatstyle{ruled}
\newfloat{listing}{tb}{lst}{}
\floatname{listing}{Listing}
\nocopyright 

\title{Enhancing Spectral Graph Neural Networks with LLM-Predicted Homophily}

\author{
    Kangkang Lu \textsuperscript{\rm 1}, Yanhua Yu \textsuperscript{\rm 1}, Zhiyong Huang \textsuperscript{\rm 2}, Tat-Seng Chua\textsuperscript{\rm 2}
}
\affiliations{
      \textsuperscript{\rm 1}Beijing University of Posts and Telecommunications \\ \textsuperscript{\rm 2}National University of Singapore
     \\ 

    lukangkang@bupt.edu.cn,   yuyanhua@bupt.edu.cn, dcshuang@nus.edu.sg, dcscts@nus.edu.sg
}

\usepackage{bibentry}

\begin{document}

\maketitle

\begin{abstract}
Spectral Graph Neural Networks (SGNNs) have achieved remarkable performance in tasks such as node classification due to their ability to learn flexible filters. Typically, these filters are learned under the supervision of downstream tasks, enabling SGNNs to adapt to diverse structural patterns. However, in scenarios with limited labeled data, SGNNs often struggle to capture the optimal filter shapes, resulting in degraded performance, especially on graphs with heterophily. Meanwhile, the rapid progress of Large Language Models (LLMs) has opened new possibilities for enhancing graph learning without modifying graph structure or requiring task-specific training. In this work, we propose a novel framework that leverages LLMs to estimate the homophily level of a graph and uses this global structural prior to guide the construction of spectral filters. Specifically, we design a lightweight and plug-and-play pipeline where a small set of labeled node pairs is formatted as natural language prompts for the LLM, which then predicts the graph's homophily ratio. This estimated value informs the spectral filter basis, enabling SGNNs to adapt more effectively to both homophilic and heterophilic structures. Extensive experiments on multiple benchmark datasets demonstrate that our LLM-assisted spectral framework consistently improves performance over strong SGNN baselines. Importantly, this enhancement incurs negligible computational and monetary cost, making it a practical solution for real-world graph applications.
\end{abstract}

\input{1_intro}
\input{2_preliminaries}

\input{3_method}

\input{4_experiment}
\input{6_conclusion}


\bibliography{aaai2026}
\newpage
\appendix
\input{7_appendix}

\end{document}

%% file: 1_intro.tex
\section{Introduction}
Graph Neural Networks (GNNs) have demonstrated impressive performance across a wide range of applications, including traffic forecasting~\cite{traffic_network_1,traffic_network_2}, anomaly detection~\cite{SmoothGNN,tang2024gadbench}, and recommendation systems~\cite{graph_recommendation1,graph_recommendation2}. By encoding both structural and attribute information into low-dimensional representations, GNNs enable effective learning on non-Euclidean data. Existing architectures can be broadly classified into spatial and spectral models. Spatial GNNs propagate messages between neighboring nodes and typically assume homophily—i.e., connected nodes share similar labels. This assumption, however, often breaks down in heterophilic graphs, where neighbors may hold dissimilar labels, leading to performance degradation~\cite{mahomophily,Hp-gmn}.

In contrast, spectral GNNs operate in the frequency domain, applying polynomial filters to modulate graph signals based on the spectrum of the Laplacian matrix. This design enables fine-grained control over low- and high-frequency information, offering improved adaptability to both homophilic and heterophilic structures~\cite{GPR-GNN,BernNet,JacobiConv}. In particular, models with learnable filters have shown strong potential in capturing complex structural patterns.

However, recent studies~\cite{NewtonNet,UniFilter} have pointed out that the effectiveness of spectral GNNs heavily depends on the quality and shape of their learned filters, which are typically supervised by downstream node classification tasks. In practice, acquiring sufficient labeled nodes is both time-consuming and expensive, especially in large-scale or sensitive domains such as biology or cybersecurity. As a result, models trained with limited labels often struggle to learn appropriate filters, leading to suboptimal generalization.

A promising yet underexplored direction is to incorporate external prior knowledge or auxiliary estimation to guide spectral filter learning. In this context, large language models (LLMs), such as GPT-4~\cite{gpt4}, emerge as powerful candidates. These models have exhibited remarkable capabilities in semantic understanding, reasoning, and generalization across multiple domains, including natural language~\cite{llms_survey}, vision~\cite{mllm_survey}, and even structured data such as graphs~\cite{llm_graph_survey1,llm_graph_survey2}. Their ability to perform few-shot or zero-shot inference makes them particularly appealing for graph tasks with limited supervision.

While prior work has attempted to combine LLMs and GNNs, existing approaches largely fall into two categories: (i) LLM-centric, where graph information is encoded into text and processed entirely by LLMs~\cite{tang2024graphgpt,zhang2024graphtranslator}, often ignoring structural nuances; and (ii) GNN-centric, where LLMs are used to augment node features or generate edges~\cite{heharnessing,xie2023graph}, which requires fine-tuning or task-specific training. These approaches either sacrifice structural expressiveness or incur considerable computational overhead, limiting their scalability and generality.

In this work, we propose a new perspective: using LLMs not to directly process graph data, but to assist spectral GNNs by estimating a key global structural property—graph homophily. Our intuition is that LLMs, given a small set of sampled edge-label pairs in natural language format, can effectively predict the overall homophily level of the graph due to their strong generalization and semantic reasoning abilities. The predicted homophily ratio can then guide the construction of spectral filters, improving GNN adaptability in both homophilic and heterophilic settings.



To this end, we propose a lightweight and general framework that leverages LLMs to estimate the homophily ratio of a graph. Specifically, we query the LLM using a small set of sampled node pairs and their labels, and aggregate its predictions to approximate the graph-level homophily ratio. This approach is cost-efficient (less than \$0.2 per dataset), requires no expensive model fine-tuning, and can be used with any black-box LLM. We then construct heterophily-aware polynomial bases guided by the predicted homophily ratio and integrate them into existing spectral GNNs. This allows the model to adapt its filter shape to the underlying graph structure.

\textbf{Our contributions are summarized as follows:}

\begin{itemize}[leftmargin=0.5cm, itemindent=0cm]
    \item \textbf{Novel Integration of LLMs into Spectral GNNs.}  
    We are the first to explore the integration of LLMs into spectral GNNs, using LLMs to estimate graph homophily and guide spectral filtering, without requiring fine-tuning, retraining, or graph structure modification.

    \item \textbf{A General and Lightweight Enhancement Framework.}  
    We propose a plug-and-play framework compatible with various spectral GNN architectures and off-the-shelf LLMs. Our method leverages the semantic reasoning abilities of LLMs to inject global structural priors into spectral models, enhancing their adaptability to different homophily regimes.

    \item \textbf{Extensive Empirical Validation.}  
    We conduct comprehensive experiments on multiple benchmark datasets, demonstrating that our framework consistently improves the performance of strong spectral GNN baselines, while incurring minimal computational and monetary overhead.
\end{itemize}

\section{Related Work}
This section introduces related research, including spectral GNNs and LLM for Graph.

\textbf{Spectral GNNs.}
Spectral GNNs can be broadly categorized based on their filtering strategies. Early methods like GCN \cite{GCN} and APPNP \cite{PPNP} use pre-defined filters based on fixed polynomial formulations. Later approaches introduce learnable filters for greater flexibility: ChebNet \cite{Chebyshev}, GPR-GNN \cite{GPR-GNN}, and BernNet \cite{BernNet} adopt various polynomial bases with trainable coefficients; JacobiConv \cite{JacobiConv} leverages orthogonal Jacobi polynomials; ChebNetII \cite{ChebNetII} uses Chebyshev interpolation; and Specformer \cite{Specformer} introduces spectral self-attention. Recently, UniFilter \cite{UniFilter} unifies heterophilic and homophilic bases to construct a universal polynomial filter that alleviates over-smoothing and over-squashing.

\textbf{LLM for Graph}. Existing research on applying large language models (LLMs) to graph learning can be broadly categorized into GNN-centric and LLM-centric approaches. GNN-centric methods typically leverage LLMs to extract node features from raw data, and then use GNNs for downstream prediction tasks \cite{heharnessing, xie2023graph}. In contrast, LLM-centric methods integrate GNNs to enhance the performance of LLMs in graph-related tasks \cite{tang2024graphgpt, zhang2024graphtranslator}. 
However, these methods are based on the homophily assumption. Beyond these, a few studies aim to enhance GNNs by leveraging LLMs to identify meaningful or noisy edges. For instance, GraphEdit \cite{graphedit} utilizes LLMs for graph structure learning by detecting and removing noisy connections; LLM4RGNN \cite{LLM4RGNN} identifies malicious and critical edges to improve the adversarial robustness of GNNs; and LLM4HeG \cite{LLM4HeG} integrates LLMs into GNNs for heterophilic graphs through edge discrimination and adaptive edge reweighting.

Existing approaches typically tailor LLM usage to specific graph tasks or graph structural modifications. In contrast, our work introduces a general and lightweight framework that utilizes LLMs to estimate the homophily ratio for enhancing spectral GNNs, without requiring LLM fine-tuning or changes to the graph. This represents a novel integration pathway between LLMs and spectral graph learning


%% file: 2_preliminaries.tex
\section{Preliminaries}

\subsection{Spectral GNN}

Assume we are given an undirected graph $\mathcal{G}=(\mathcal{V}, \mathcal{E}, \mathbf{X})$, where $\mathcal{V}=\left\{v_1, \ldots, v_n\right\}$ denotes the set of $n$ nodes, $\mathcal{E}$ is the edge set, and $\mathbf{X} \in \mathbb{R}^{n \times d}$ is the node feature matrix. The corresponding adjacency matrix is $\mathbf{A} \in \{0,1\}^{n \times n}$, where $\mathbf{A}_{ij}=1$ if there is an edge between nodes $v_i$ and $v_j$, and $\mathbf{A}_{ij} = 0$ otherwise. The degree matrix $\mathbf{D} = \text{diag}(d_1,\ldots,d_n)$ is a diagonal matrix where the $i$-th diagonal entry is $d_i = \sum_j \mathbf{A}_{ij}$. The normalized Laplacian matrix is defined as $\mathbf{\hat{L}} = \mathbf{I} - \mathbf{D}^{-\frac{1}{2}} \mathbf{A} \mathbf{D}^{-\frac{1}{2}}$, where $\mathbf{I}$ denotes the identity matrix. The normalized adjacency matrix is $\mathbf{\hat{A}} = \mathbf{D}^{-\frac{1}{2}} \mathbf{A} \mathbf{D}^{-\frac{1}{2}}$. Let $\mathbf{\hat{L}} = \mathbf{U} \boldsymbol{\Lambda} \mathbf{U}^{\top}$ denote the eigen-decomposition of $\mathbf{\hat{L}}$, where $\mathbf{U}$ is the matrix of eigenvectors and $\boldsymbol{\Lambda} = \text{diag}([\lambda_1, \lambda_2, \ldots, \lambda_n])$ is the diagonal matrix of eigenvalues.

Spectral GNN is based on the Fourier transform in signal processing. The Fourier transform of a graph signal \(\mathbf{x}\) is given by \(\hat{\mathbf{x}} = \mathbf{U}^{\top} \mathbf{x}\), and its inverse is \(\mathbf{x} = \mathbf{U} \hat{\mathbf{x}}\). Accordingly, the graph convolution of the signal \(\mathbf{x}\) with a kernel \(\mathbf{g}\) can be defined as:

\begin{equation}
\label{eq:Fourier_transform}
\mathbf{z} = \mathbf{g} *_{\mathcal{G}} \mathbf{x} = \mathbf{U} \left( (\mathbf{U}^{\top} \mathbf{g}) \odot (\mathbf{U}^{\top} \mathbf{x}) \right) = \mathbf{U} \mathbf{\hat{G}} \mathbf{U}^{\top} \mathbf{x},
\end{equation}

where $\mathbf{\hat{G}} = \operatorname{diag}(\hat{g}_1, \ldots, \hat{g}_n)$ denotes the spectral kernel coefficients. To avoid explicit eigen-decomposition, recent works approximate different kernels $\mathbf{H}$ using polynomial functions $h(\cdot)$ as follows:

\begin{equation}
\label{eq:h_polynomial}
\mathbf{H} = h(\mathbf{\hat{L}}) = h_0 \mathbf{\hat{L}}^0 + h_1 \mathbf{\hat{L}}^1 + h_2 \mathbf{\hat{L}}^2 + \cdots + h_K \mathbf{\hat{L}}^K,
\end{equation}

where $K$ is the order of the polynomial $h(\cdot)$ and $h_k$ is the coefficient of the $k$-th order term. Thus, Eq.~(\ref{eq:Fourier_transform}) can be rewritten as:
\begin{equation}
\begin{aligned}
\mathbf{Z}  & = \mathbf{H} \mathbf{X} = h(\mathbf{\hat L}) \mathbf{X}  =\mathbf{U} h(\boldsymbol{\Lambda}) \mathbf{U}^{\top}  \mathbf{X} \\
&= \mathbf{U}\left(\begin{array}{ccc}
h(\lambda_1) & \cdots & 0 \\
\vdots & \ddots & \vdots \\
0 & \cdots & h(\lambda_n)
\end{array}\right) \mathbf{U}^{\top} \mathbf{X},
\label{eq:filter}
\end{aligned}
\end{equation}

where $\mathbf{Z}$ is the output (prediction) matrix.

\subsection{Homophily}


The homophily metric measures the degree of association between connected nodes. Homophily can be measured in various ways, such as node-level homophily \cite{node_homo}, edge-level homophily \cite{edge_homo}, and class-level homophily \cite{class_homo}. The widely adopted edge homophily is defined as follows:
\begin{equation}
\label{eq:homophily_edge}	
\mathcal{H}_{\text {edge }}(\mathcal{G})=\frac{1}{|\mathcal{E}|} \sum_{(u, v) \in \mathcal{E}} \mathbf{1} \left(y_u=y_v\right),
\end{equation}
where $\mathbf{1} (\cdot)$ is the indicator function, i.e., $\mathbf{1} (\cdot)=1$ if the condition holds, otherwise $\mathbf{1} (\cdot)=0$. $y_u$ and $y_v$ denote the labels of nodes $u$ and $v$, respectively. $|\mathcal{E}|$ is the size of the edge set.

%% file: 3_method.tex
\section{Methodology}

\begin{figure*}[t] 
	\includegraphics[width=1\linewidth]{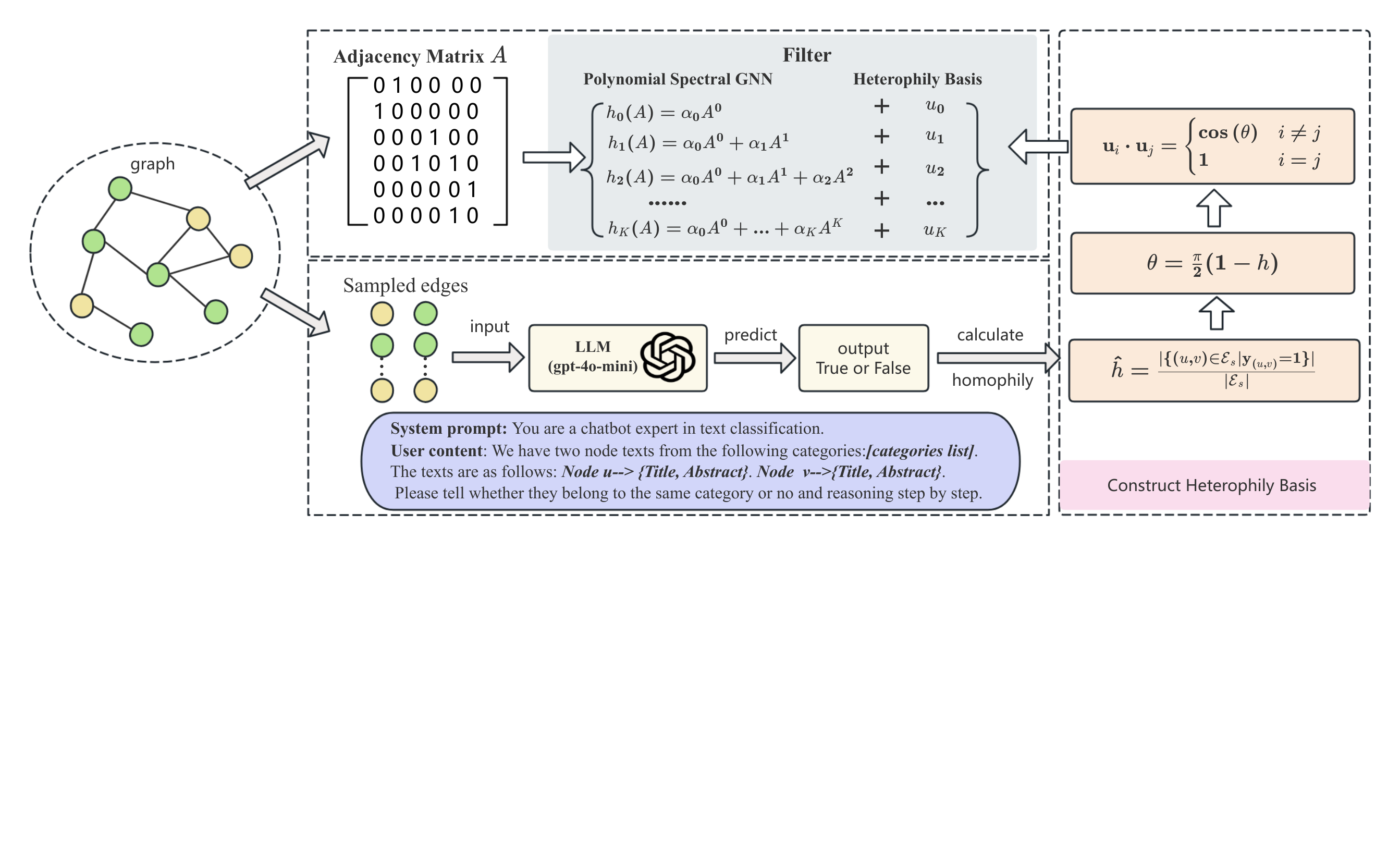}
	\caption{The overall framework of the proposed model. First, we sample a subset of edges (node pairs) from the dataset. Then we use LLM to predict whether the connected nodes belong to the same class. Based on the LLM's predictions, we estimate the graph's homophily. Finally, a heterophily basis is constructed using the predicted homophily and integrated into existing polynomial spectral GNNs to perform the node classification task.}

	\label{fig:model}
\end{figure*}

This section describes the proposed method, which consists of two modules: (1) using an LLM to predict the homophily ratio, and (2) incorporating the predicted homophily into existing spectral GNNs. The following provides a detailed introduction to these two modules.

\subsection{Estimating Homophily Using LLM}

It is well established that the homophily ratio of a graph quantifies the degree of label similarity among neighboring nodes, reflecting the proportion of edges that connect nodes belonging to the same class. Consequently, accurate estimation of this ratio is essential for leveraging structural properties of the graph and enabling effective spectral filtering. For instance, in highly homophilic graphs, a polynomial spectral GNN is expected to learn a low-pass filter~\cite{FAGCN}.


\begin{table}[!ht]
    \centering
    \caption{Comparison of homophily predicted by the LLM with varying prompts, with indicators for whether CoT and Most-voting (Most) are used. The percentage reflects the closeness to ground truth (higher is better).}
    \begingroup
    \setlength{\tabcolsep}{3pt}
    \fontsize{9pt}{10pt}\selectfont
    \begin{tabular}{cc|ccc|c}
        \toprule
       \multicolumn{2}{c|}{Prompt Type} & \multirow{2}{*}{ Cora} & \multirow{2}{*}{ Citeseer} & \multirow{2}{*}{Pubmed} & \multirow{2}{*}{Average} \\
        \cline{1-2}
        CoT & Most & & & & \\
        \midrule
        × & × & 0.41(49\%) & 0.63(81\%) & 0.72(90\%) & 73\% \\
        \checkmark & × & 0.51(63\%) & 0.76(97\%) & 0.67(84\%) & 81\% \\
        × & \checkmark & 0.37(46\%) & 0.63(81\%) & 0.81(99\%) & 75\% \\
        \checkmark & \checkmark & 0.70(86\%) & 0.81(96\%) & 0.83(98\%) & 93\% \\
        \midrule
        \multicolumn{2}{c|}{Ground Truth} & 0.81 & 0.78 & 0.80 & -- \\
        \bottomrule
    \end{tabular}
    \endgroup
\end{table}

For simplicity, we adopt the widely used edge-level homophily \cite{edge_homo} as the target to estimate. Specifically, we aim to determine the proportion of homophilic edges—edges connecting nodes of the same class—among all edges. However, exhaustively evaluating every edge is impractical, especially for large-scale graphs with millions of edges, due to the high inference cost of LLMs. Thus, we estimate the overall homophily by sampling a representative subset of edges.


To ensure accurate estimation of homophily, it is important to improve the quality of individual edge predictions produced by the LLM. Motivated by recent advances~\cite{LLM-GNN} in enhancing prediction reliability via prompt design, we explore the following strategies:

\begin{itemize}[leftmargin=0.5cm, itemindent=0cm]
\item \textbf{Vanilla prompting:} Directly querying the model without any additional reasoning or consistency mechanisms.
\item \textbf{Reasoning-based prompting (CoT):} Encouraging the model to perform step-by-step reasoning using chain-of-thought prompts.
\item \textbf{Consistency-based prompting (Most-voting):} Querying the model multiple times and using majority voting to determine the final prediction.
\item \textbf{Hybrid prompting:} Combining both reasoning and consistency to improve robustness and accuracy.
\end{itemize}


As shown in Table~1, both reasoning and consistency-based prompting enhance homophily prediction accuracy. Therefore, we adopt a hybrid prompting strategy to combine their complementary strengths. The following is an example prompt:


\begin{tcolorbox}[top=2pt, bottom=2pt, left=2pt, right=2pt]
\textbf{System:} You are a chatbot expert in text classification.\\
\textbf{User:} We have two node texts from the following categories: \textit{[categories list]}. The texts are as follows: \\
\textbf{\textit{ Node $v_i$$\rightarrow$\{Title,Abstract\}. Node $v_j$$\rightarrow$\{Title,Abstract\}.}} \\
Please tell me whether they belong to the same category or not after reasoning step by step.
\end{tcolorbox}


After obtaining initial predictions using reasoning-based prompting, we further enhance reliability via consistency-based voting. Specifically, each sampled edge is queried five times, and the final label is determined via majority voting. Let $y_{(u, v)} \in \{0, 1\}$ denote the final label for edge $(u, v)$, where 1 indicates nodes $u$ and $v$ belong to the same class (i.e. homophilic edge), and 0 otherwise. The decision rule is defined as:

\[
y_{(u, v)}= \left\{
\begin{array}{ll}
1 & \text{if } r_{(u, v)} \geq 3, \\
0 & \text{if } r_{(u, v)} < 3,
\end{array}
\right.
\]


where $r_{(u, v)}$ denotes the number of responses labeling edge $(u, v)$ as homophilic among the five queries. Finally, the edge-level homophily $\hat{h}$ is estimated as:

\begin{equation}
\hat{h} = \frac{|\{ (u, v) \in \mathcal{E}_s \mid y_{(u, v)} = 1 \}|}{|\mathcal{E}_s|},
\end{equation}

where $\mathcal{E}_s$ denotes the set of all sampled edges.



\subsection{Incorporating Homophily into Polynomial Spectral Filters}

After estimating the homophily ratio using an LLM, we integrate the predicted homophily into a polynomial spectral GNN framework. To enable spectral GNNs to exploit the homophily information inferred in Eq.~(5), we construct a set of heterophily-aware basis vectors, inspired by UniFilter~\cite{UniFilter}. Specifically, the angle between each pair of basis vectors is defined as:
$$
\theta = \frac{\pi}{2}(1 - \hat{h}).
$$

This design is motivated by theoretical findings in UniFilter \cite{UniFilter}, which suggest that the angular separation between basis vectors is positively correlated with their spectral frequency. Since ideal signal frequency is linearly proportional to $(1 - h)$, setting $\theta = \frac{\pi}{2}(1 - \hat{h})$ allows the constructed basis to adaptively reflect the graph's degree of heterophily, thus mitigating over-smoothing in low-homophily settings. Thus, the angle between different heterophily-aware basis vectors \( \mathbf{u}_i \) and \( \mathbf{u}_j \) is given by:

$$
\mathbf{u}_i \cdot \mathbf{u}_j=  \begin{cases}\cos \theta =  \cos \left(\frac{(1-\hat h) \pi}{2}\right) & \text { if } i \neq j, \\ 1 & \text { if } i=j.\end{cases}
$$

After obtaining the heterophily basis vectors set \( \left\{\mathbf{u}_0, \mathbf{u}_1, \cdots, \mathbf{u}_K\right\} \), we incorporate them into existing polynomial spectral GNNs, such as  GPRGNN~\cite{GPR-GNN}, BernNet~\cite{BernNet}, JacobiConv~\cite{JacobiConv}, and ChebNetII~\cite{ChebNetII}. The details are as follows:

\textbf{Insertion into GPR-GNN.}
 GPR-GNN \cite{GPR-GNN} directly assigns a learnable coefficient to each order of the normalized adjacency matrix $\mathbf{\hat A}$, and its polynomial filter is defined as:
\begin{equation}
\mathbf{z}=\sum_{k=0}^K \gamma_k  \mathbf{\hat A}^k = \mathbf{U} g_{\gamma, K}(\Lambda) \mathbf{U}^T ,
\end{equation}
where $g_{\gamma, K}(\Lambda)$ is an element-wise operation, and $g_{\gamma, K}(x)=\sum_{k=0}^K \gamma_k x^k$. GPR-GNN represents the simplest form of polynomial spectral GNN, assigning a single scalar coefficient to each propagation step. Therefore, we directly incorporate heterophily basis vectors into GPR-GNN in ascending order of their polynomial order:

\begin{equation}
\mathbf{z}=\sum_{k=0}^K \gamma_k \left(\beta \mathbf{\hat A}^k \mathbf{x}+(1-\beta) \mathbf{u}_k\right),
\end{equation}
where $\beta$ is a tunable hyperparameter.

\begin{table*}[t]
    \centering
    \caption{Dataset Statistics}
    \label{tab:datasets}
    \begin{tabular}{lcccccccccc}
        \toprule
       Dataset & Washington  & Cornell & Texas & Children & History & Fitness & Cora & Citeseer & Pubmed \\
        \midrule
       Nodes & 229  & 191 & 187 & 76,875 & 41,551 & 173,055 & 2,708 & 3,186 & 19,717 \\
        Edges & 394  & 292 & 310 & 1,554,578 & 358,574 & 1,773,500 & 5,429 & 4,277 & 44,335 \\ 
        Feat & 768  & 768 & 768 & 768 & 768 & 768 & 1,433 & 3,703 & 500 \\ 
        Classes & 5  & 5 & 5 & 24 & 12 & 13 & 7 & 6 & 3 \\ 
        Homophily & 0.15  & 0.12 & 0.06 & 0.42 & 0.66 & 0.90 & 0.81 & 0.78 & 0.80 \\
        \bottomrule
    \end{tabular}
\end{table*}

\textbf{Insertion into BernNet.}
BernNet \cite{BernNet} expresses the filtering operation with Bernstein polynomials and forces all coefficients to be positive, and its filter is defined as:
\begin{equation}
\mathbf{z}=\sum_{k=0}^K \theta_k \frac{1}{2^K}\left(\begin{array}{c}
K \\
k
\end{array}\right)(2 \mathbf{I}-\mathbf{L})^{K-k} \mathbf{L}^k  \mathbf{x}.
\end{equation}
For BernNet, each term is a product involving both \( 2\mathbf{I} - \mathbf{L} \) and \( \mathbf{L} \). We insert the \( k \)-th heterophily basis vector \( \mathbf{u}_k \) into the \( k \)-th order of \( \mathbf{L} \):
\begin{equation}
\mathbf{z}=\sum_{k=0}^K \theta_k [\beta \frac{1}{2^K}\left(\begin{array}{c}
K \\
k
\end{array}\right)(2 \mathbf{I}-\mathbf{L})^{K-k} \mathbf{L}^k  \mathbf{x} +(1-\beta) \mathbf{u}_k ].
\end{equation}


\textbf{Insertion into JacobiConv.}
JacobiConv \cite{JacobiConv} proposes a Jacobi basis to adapt a wide range of weight functions due to its orthogonality and flexibility. The iterative process of the Jacobi basis can be defined as:
\begin{equation}
\begin{aligned}
& P_0^{a, b}(x)=1 , \\
& P_1^{a, b}(x)=0.5 a-0.5 b+(0.5 a+0.5 b+1) x , \\
& P_k^{a, b}(x)=(2 k+a+b-1) \\
& \cdot \frac{(2 k+a+b)(2 k+a+b-2) x +a^2-b^2}{2 k(k+a+b)(2 k+a+b-2)} P_{k-1}^{a, b}(x) \\
& -\frac{(k+a-1)(k+b-1)(2 k+a+b)}{k(k+a+b)(2 k+a+b-2)} P_{k-2}^{a, b}(x) ,
\end{aligned}
\end{equation}
where $a$ and $b$ are tunable hyperparameters. Unlike GPR-GNN and BernNet, JacobiConv adopts an individual filter function for each output dimension $l$:       
\begin{equation}
\mathbf{Z}_{: l}=\sum_{k=0}^K \alpha_{k l} P_k^{a, b}(\mathbf{\hat A}) (\mathbf{X}\mathbf{W})_{: l} .
\end{equation}
Similarly, we incorporate the corresponding heterophily basis vector $\mathbf{u}_k$ into each polynomial order of JacobiConv as follows:

\begin{equation}
\mathbf{Z}_{: l}=\sum_{k=0}^K \alpha_{k l} [ \beta P_k^{a, b}(\mathbf{\hat A}) (\mathbf{X}\mathbf{W})_{: l} +(1-\beta) \mathbf{u}_k ].
\end{equation}

\textbf{Insertion Details of ChebNetII.}
To mitigate overfitting and reduce the Runge phenomenon, ChebNetII \cite{ChebNetII} enhances the original Chebyshev  \cite{Chebyshev} polynomial approximation by reparameterizing the filter response as a learnable coefficient $\gamma_j$ , as shown below:

\begin{equation}
\mathbf{z}=\frac{2}{K+1} \sum_{k=0}^K \sum_{j=0}^K \gamma_j T_k\left(x_j\right) T_k(\mathbf{L}) \mathbf{x},
\end{equation}

where $x_j=\cos ((j+1 / 2) \pi /(K+1))$ are the Chebyshev nodes of $T_{K+1}$. The recursive definition of the Chebyshev polynomial $T_{k}(x)$ is given as follows:

\begin{equation}
\begin{aligned}
T_0(x) &= 1, \\
T_1(x)& = x, \\
T_k(x)&  = 2xT_{k-1}(x) - T_{k-2}(x).
\end{aligned}
\end{equation}

We incorporate the heterophily basis vector \( \mathbf{u}_k \) into the corresponding \( k \)-th term of the ChebNetII, resulting in:

\begin{equation}
\mathbf{z}=\frac{2}{K+1} \sum_{k=0}^K \left[ \beta \sum_{j=0}^K \gamma_j T_k\left(x_j\right) T_k(\mathbf{L}) \mathbf{x} + (1 - \beta)\mathbf{u}_k\right] ,
\end{equation}

\subsection{Training Objective}
After integrating the LLM-estimated homophily ratio into existing polynomial spectral filters, node classification can be performed using various polynomial-based spectral GNNs. Notably, the proposed method introduces no additional trainable parameters.


We adopt a multi-layer perceptron (MLP) with parameter $\theta$ to predict the label distribution:
\begin{equation}
\hat{\mathbf{y}}=\operatorname{MLP}\left(\mathbf{Z} ; \theta\right),
\end{equation}
where $\hat{\mathbf{y}}$ is the predicted label distribution. Then, we optimize the cross-entropy loss of the node $j$:
\begin{equation}
\mathcal{L}=\sum_{j \in \mathcal{V}_{\text {train }}} \operatorname{CrossEntropy}\left(\hat{\mathbf{y}}^j, \mathbf{y}^j\right),
\end{equation}
where $\mathcal{V}_{\text {train }}$ is the training node set, and $\mathbf{y}^j$
is the ground-truth one-hot label vector of node $j$.

%% file: 4_experiment.tex
\section{Experiment}

In this section, to fully evaluate the performance of the proposed model, we present a series of comprehensive experiments to answer the following research questions (\textbf{RQ}s): 

\begin{itemize}[leftmargin=0.5cm, itemindent=0cm]
\item \textbf{RQ1}: How does the proposed method perform compared to its corresponding spectral GNN counterparts?
\item \textbf{RQ2}: How effective is LLM-predicted homophily compared to that estimated by MLPs or directly derived from the training set?
\item \textbf{RQ3}: Does the LLM-based homophily estimation incur negligible time and money cost?
\item \textbf{RQ4}: How does the key hyperparameter $\beta$ influence the model’s performance?
\end{itemize}


\subsection{Experimental Setup}
\textbf{Datasets.}
We select nine graph datasets with text attributes \cite{CS-TAG, graphedit}
, including three citation networks (Cora, Citeseer, and Pubmed), three webpage networks (Cornell, Texas, and Washington), and three Amazon co-purchase networks (Children, History, and Fitness). The statistics of these datasets are summarized in Table \ref{tab:datasets}.

\begin{table*}[!ht]
    \centering
    \caption{Performance (\%) on Various Datasets (mean accuracy with standard deviation as subscript)}
    \label{tab:results}
     \begingroup
    \setlength{\tabcolsep}{3pt}
    \fontsize{9pt}{11pt}\selectfont
    \begin{tabular}{lcccccccccc}
        \toprule
        Method & Washington & Cornell & Texas & Children & History & Fitness & Cora & Citeseer & Pubmed \\
        \midrule
        MLP & 80.85$_{\pm0.00}$ & 68.38$_{\pm1.48}$ & 79.82$_{\pm1.52}$ & 54.90$_{\pm0.04}$ & 83.91$_{\pm0.14}$ & 79.57$_{\pm0.06}$ & 79.52$_{\pm0.32}$ & 71.68$_{\pm1.68}$ & 87.43$_{\pm0.11}$ \\ 
        GCN & 57.45$_{\pm4.26}$ & 37.61$_{\pm10.36}$ & 57.02$_{\pm1.52}$ & 56.47$_{\pm0.10}$ & 84.82$_{\pm0.05}$ & 90.86$_{\pm0.06}$ & 86.22$_{\pm0.57}$ & 78.58$_{\pm0.81}$ & 85.02$_{\pm0.21}$ \\
        GraphSAGE & 78.72$_{\pm2.13}$ & 69.23$_{\pm4.44}$ & 76.32$_{\pm2.64}$ & 54.39$_{\pm1.32}$ & 84.51$_{\pm0.19}$ & 88.55$_{\pm2.07}$ & 86.65$_{\pm0.77}$ & 76.96$_{\pm1.36}$ & 88.80$_{\pm0.31}$ \\
        GAT & 44.68$_{\pm11.85}$ & 30.77$_{\pm10.26}$ & 51.75$_{\pm15.20}$ & 49.78$_{\pm4.31}$ & 82.79$_{\pm1.18}$ & 90.51$_{\pm0.68}$ & 84.93$_{\pm0.76}$ & 78.16$_{\pm0.45}$ & 85.04$_{\pm0.38}$ \\
        APPNP & 68.79$_{\pm2.45}$ & 57.26$_{\pm1.48}$ & 62.28$_{\pm1.52}$ & 45.34$_{\pm0.27}$ & 81.42$_{\pm0.07}$ & 80.33$_{\pm0.36}$ & 88.01$_{\pm0.64}$ & 81.40$_{\pm0.33}$ & 87.58$_{\pm0.03}$ \\
        TFEGNN & 68.09$_{\pm9.19}$ & 66.67$_{\pm5.54}$ & 64.04$_{\pm1.24}$ & 45.44$_{\pm0.83}$ & 81.19$_{\pm0.03}$ & 83.31$_{\pm1.02}$ & 85.50$_{\pm0.27}$ & 77.83$_{\pm0.52}$ & 88.25$_{\pm0.17}$ \\
        UniFilter & 88.61$_{\pm4.16}$ & 69.00$_{\pm4.12}$ & 79.38$_{\pm3.65}$ & 47.55$_{\pm0.16}$ & 86.99$_{\pm0.84}$ & 87.81$_{\pm4.74}$ & 87.19$_{\pm1.73}$ & 77.57$_{\pm2.10}$ & 85.55$_{\pm3.63}$ \\
        \midrule
        GPRGNN & 82.98$_{\pm2.13}$ & 68.38$_{\pm2.96}$ & 81.58$_{\pm2.63}$ & 60.29$_{\pm0.46}$ & 86.03$_{\pm0.07}$ & 92.38$_{\pm0.07}$ & 88.01$_{\pm0.19}$ & 78.16$_{\pm0.33}$ & 89.11$_{\pm0.19}$ \\
        \textbf{GPRGNNPLUS} & \textbf{87.23$_{\pm0.00}$} & \textbf{70.94$_{\pm2.96}$} & \textbf{83.33$_{\pm1.52}$} & \textbf{61.00$_{\pm0.20}$} & \textbf{86.47$_{\pm0.13}$} & \textbf{92.58$_{\pm0.04}$} & \textbf{88.07$_{\pm0.21}$} & \textbf{79.94$_{\pm0.41}$} & \textbf{89.53$_{\pm0.09}$} \\
        \midrule
        BernNet & 79.43$_{\pm1.23}$ & 67.52$_{\pm1.48}$ & 80.70$_{\pm1.52}$ & 58.76$_{\pm0.17}$ & 84.93$_{\pm0.30}$ & 91.83$_{\pm0.16}$ & 82.41$_{\pm0.60}$ & 72.83$_{\pm1.48}$ & 88.54$_{\pm0.36}$ \\
        \textbf{BernNetPLUS} & \textbf{90.78$_{\pm1.23}$} & \textbf{71.79$_{\pm5.13}$} & \textbf{83.33$_{\pm1.52}$} & \textbf{60.80$_{\pm0.19}$} & \textbf{85.66$_{\pm0.09}$} & \textbf{92.44$_{\pm0.06}$} & \textbf{87.58$_{\pm1.06}$} & \textbf{78.89$_{\pm1.28}$} & \textbf{89.37$_{\pm0.08}$} \\
        \midrule
        JacobiConv & 78.01$_{\pm2.45}$ & 64.96$_{\pm3.92}$ & 79.82$_{\pm3.04}$ & 59.63$_{\pm0.46}$ & 85.36$_{\pm0.13}$ & 90.70$_{\pm0.17}$ & 84.99$_{\pm1.01}$ & 72.88$_{\pm1.13}$ & 88.04$_{\pm0.10}$ \\
        \textbf{JacobiConvPLUS} & \textbf{80.85$_{\pm0.00}$} & \textbf{68.38$_{\pm1.48}$} & \textbf{80.70$_{\pm1.52}$} & \textbf{59.67$_{\pm0.05}$} & \textbf{85.79$_{\pm0.07}$} & \textbf{91.04$_{\pm0.26}$} & \textbf{88.25$_{\pm0.39}$} & \textbf{78.63$_{\pm1.48}$} & \textbf{89.48$_{\pm0.13}$} \\
        \midrule
        ChebNetII & 82.27$_{\pm1.00}$ & 68.38$_{\pm1.21}$ & 84.21$_{\pm2.15}$ & 56.74$_{\pm0.07}$ & 84.76$_{\pm0.10}$ & 83.49$_{\pm0.02}$ & 81.01$_{\pm0.12}$ & 72.37$_{\pm0.24}$ & 88.34$_{\pm0.13}$ \\
        \textbf{ChebNetIIPLUS} & \textbf{87.94$_{\pm1.00}$} & \textbf{69.23$_{\pm2.09}$} & \textbf{84.21$_{\pm2.15}$} & \textbf{58.76$_{\pm0.12}$} & \textbf{85.79$_{\pm0.05}$} & \textbf{88.22$_{\pm0.17}$} & \textbf{83.53$_{\pm0.17}$} & \textbf{74.12$_{\pm0.36}$} & \textbf{88.34$_{\pm0.10}$} \\
        \bottomrule
    \end{tabular}
    \endgroup
\end{table*}

\textbf{Settings.} 
We adopt the experimental setup used in CS-TAG~\cite{CS-TAG} and follow its standard data split to ensure a fair comparison. Specifically, the training/validation/test sets are divided as 60\%/20\%/20\% for all datasets except Fitness, for which the split is 20\%/10\%/70\%. We use the accuracy metric as an evaluation indicator. All experiments are performed three times, and we report the average results and their corresponding standard errors. All experiments are conducted on a machine with 3 NVIDIA A5000 24GB GPUs and Intel(R) Xeon(R) Silver 4310 2.10 GHz CPU. The LLM used for generating homophily ratio predictions is GPT-4o-mini \cite{Gpt-4o}.

\textbf{Baselines.}
To thoroughly evaluate the effectiveness of the proposed method, in addition to the four polynomial spectral GNN backbones---GPR-GNN~\cite{GPR-GNN}, BernNet~\cite{BernNet}, JacobiConv~\cite {JacobiConv}, and ChebNetII~\cite{ChebNetII}---we also include six classical GNN models and a Multi-Layer Perceptron (MLP) as baselines: GCN~\cite{GCN}, GAT~\cite{GAT}, GraphSAGE~\cite {GraphSAGE}, APPNP~\cite {PPNP}, TFE-GNN~\cite{TFEGNN}, and UniFilter~\cite {UniFilter}.

For more experimental details, including comprehensive descriptions of the baselines, datasets, and hyperparameter settings, please refer to Appendix~A.

\subsection{Main Results (\textbf{RQ1})}
Table~\ref{tab:results} reports the node classification accuracies on nine benchmark datasets. As observed, our LLM-enhanced variants (denoted with the \textbf{PLUS} suffix) consistently outperform their original counterparts, including GPRGNN, BernNet, JacobiConv, and ChebNetII, across all datasets. Concretely, the average performance gains over the original models are substantial: \textbf{+1.17\%} for GPRGNN, \textbf{+4.93\%} for BernNet, \textbf{+2.70\%} for JacobiConv, and \textbf{+2.68\%} for ChebNetII. These consistent improvements clearly demonstrate the effectiveness of incorporating LLM-predicted homophily into spectral GNNs.

In addition, despite the competitive performance of recent spectral methods such as UniFilter \cite{UniFilter} and TFE-GNN \cite{TFEGNN}, our approach still achieves superior results across diverse datasets. Importantly, our method introduces no graph topology modifications or task-specific training, yet enhances the adaptability of spectral GNNs to different structural patterns, including both homophilic and heterophilic graphs.

Overall, the results demonstrate that incorporating global structural priors via LLMs can significantly boost the representational power of spectral GNNs, offering a simple yet effective enhancement to existing models.



\begin{figure}[t]   
	\centering 
	\includegraphics[width=\linewidth]{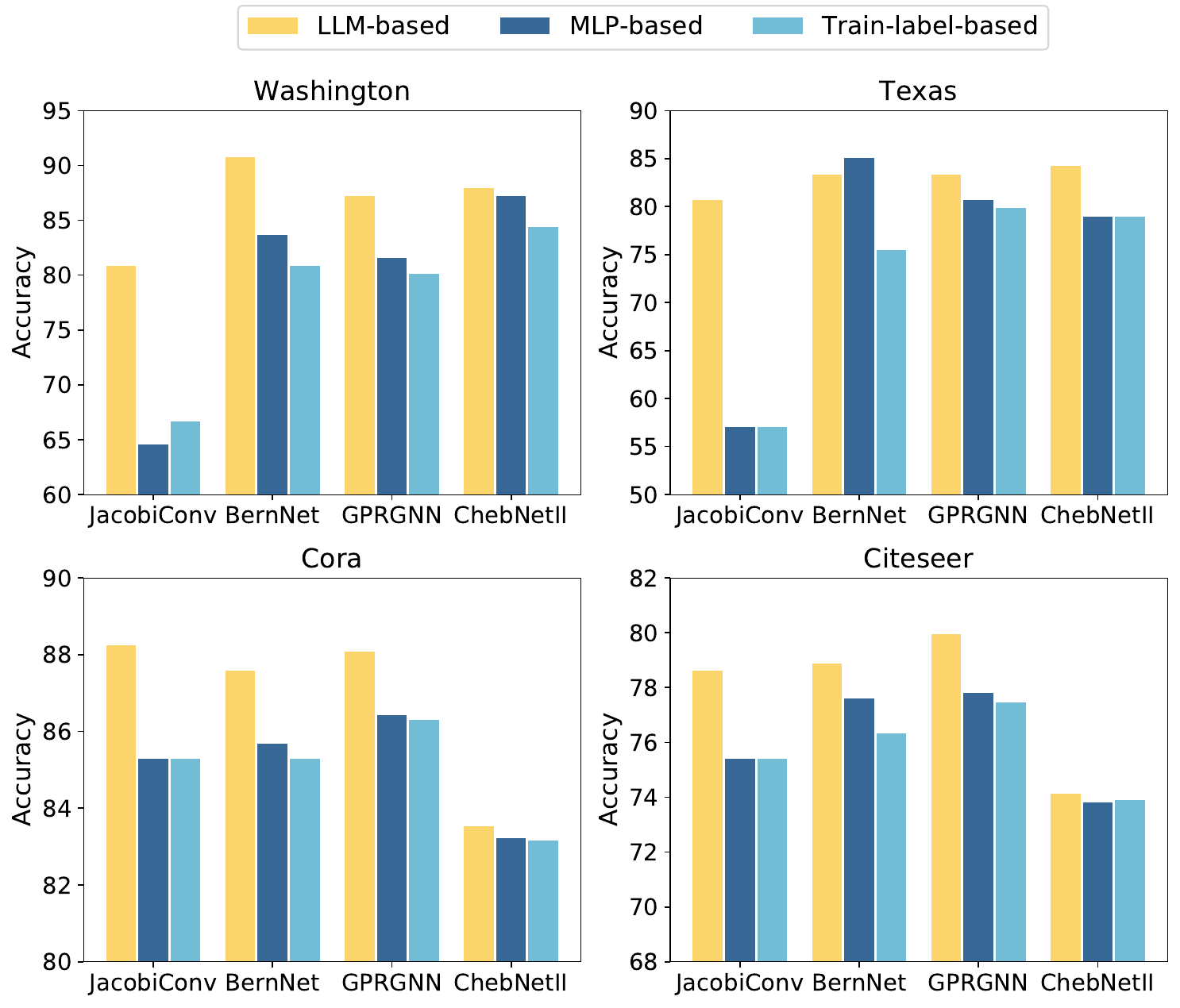}
	\caption{Ablation study of proposed method.}
 \label{fig:ablation}
\end{figure}




\subsection{Ablation Analysis (\textbf{RQ2})}
We evaluate the impact of incorporating LLM-predicted homophily into spectral GNNs (\textbf{LLM-based}) by comparing it with two variants: 
(1) \textbf{MLP-based}, which trains an MLP to classify each edge as homophilic or heterophilic for estimating overall homophily; 
(2) \textbf{Train-label-based}, which directly computes homophily from the proportion of homophilic edges among labeled node pairs in the training set.
Figure~\ref{fig:ablation} presents the performance of these variants on four datasets. As shown, LLM-predicted homophily consistently achieves the best results, with the only exception being BernNet on the Texas dataset. This underscores the advantage of LLMs in generating more reliable structural priors, likely due to their stronger reasoning capabilities. Notably, MLP-based estimation also outperforms the direct use of training-set homophily, indicating that learning homophily signals is more effective than relying on limited labeled data. However, it still lags behind the LLM-based variant, highlighting that the quality of the estimated homophily plays a crucial role in guiding spectral filter construction.

\begin{table*}[!ht]
    \centering
    \caption{Token and cost statistics on each dataset (million tokens / USD)}
     \begingroup
    \fontsize{9pt}{11pt}\selectfont
    \begin{tabular}{lccccccccc}
        \toprule
        Type & Washington & Cornell & Texas & Children & History & Fitness & Cora & Citeseer & Pubmed \\
        \midrule
        Input  & 0.47/0.07\$ & 0.56/0.08\$ & 0.44/0.07\$ & 0.38/0.06\$ & 0.36/0.05\$ & 0.09/0.01\$ & 0.22/0.03\$ & 0.24/0.04\$ & 0.40/0.06\$ \\
        Output & 0.12/0.07\$ & 0.14/0.09\$ & 0.13/0.08\$ & 0.17/0.10\$ & 0.16/0.10\$ & 0.15/0.09\$ & 0.17/0.10\$ & 0.16/0.10\$ & 0.18/0.11\$ \\
        Total  & 0.59/0.14\$ & 0.71/0.17\$ & 0.57/0.14\$ & 0.55/0.16\$ & 0.53/0.15\$ & 0.24/0.10\$ & 0.39/0.13\$ & 0.40/0.13\$ & 0.58/0.17\$ \\
        \bottomrule
    \end{tabular}
    \label{tab:cost}
    \endgroup
\end{table*}

\begin{table*}[!ht]
    \centering
    \caption{Per-epoch training time (ms) and total training time (s) comparisons on various datasets.}
    \label{tab:time}
    \begingroup
    \setlength{\tabcolsep}{4pt}
    \fontsize{9pt}{10pt}\selectfont
    \begin{tabular}{lccccccccc}
        \toprule
        Method & Washington & Cornell & Texas & Children & History & Fitness & Cora & Citeseer & Pubmed \\
        \midrule
        GPRGNN & 13.0/2.6 & 11.8/2.4 & 13.4/2.7 & 230.8/46.2 & 94.9/19.0 & 472.7/94.5 & 15.0/3.0 & 31.4/6.3 & 32.6/6.5 \\
        \textbf{GPRGNNPLUS} & 13.6/2.7 & 13.9/2.8 & 13.9/2.8 & 274.0/54.8 & 119.1/23.8 & 575.4/115.1 & 18.8/3.8 & 41.4/8.3 & 40.6/8.1 \\
        \midrule
        BernNet & 37.5/7.5 & 38.6/7.7 & 35.9/7.2 & 1235.3/247.1 & 445.7/89.2 & 2449.4/489.9 & 50.2/10.0 & 132.1/26.4 & 128.3/25.7 \\
        \textbf{BernNetPLUS} & 43.5/8.7 & 43.8/8.8 & 43.5/8.7 & 1303.5/260.7 & 473.8/94.8 & 2596.2/519.2 & 53.7/10.7 & 143.1/28.6 & 138.2/27.7 \\
        \midrule
        JacobiConv & 26.8/5.4 & 26.8/5.4 & 26.8/5.4 & 56.8/11.4 & 33.9/6.8 & 88.0/17.6 & 28.8/5.8 & 28.1/5.6 & 30.1/6.0 \\
        \textbf{JacobiConvPLUS} & 28.1/5.6 & 27.9/5.6 & 28.0/5.6 & 56.5/11.3 & 34.4/6.9 & 86.3/17.3 & 30.5/6.1 & 29.8/6.0 & 31.3/6.3 \\
        \midrule
        ChebNetII & 49.4/9.9 & 49.6/9.9 & 49.2/9.8 & 274.7/55.0 & 137.0/27.4 & 574.5/114.9 & 51.8/10.4 & 70.4/14.1 & 68.8/13.8 \\
        \textbf{ChebNetIIPLUS} & 50.0/10.0 & 50.6/10.1 & 51.2/10.2 & 324.2/64.8 & 164.5/32.9 & 685.7/137.1 & 54.8/11.0 & 79.5/15.9 & 76.9/15.4 \\
        \bottomrule
    \end{tabular}
    \endgroup
\end{table*}

\begin{figure}[t]   
	\centering 
 
	\includegraphics[width=\linewidth]{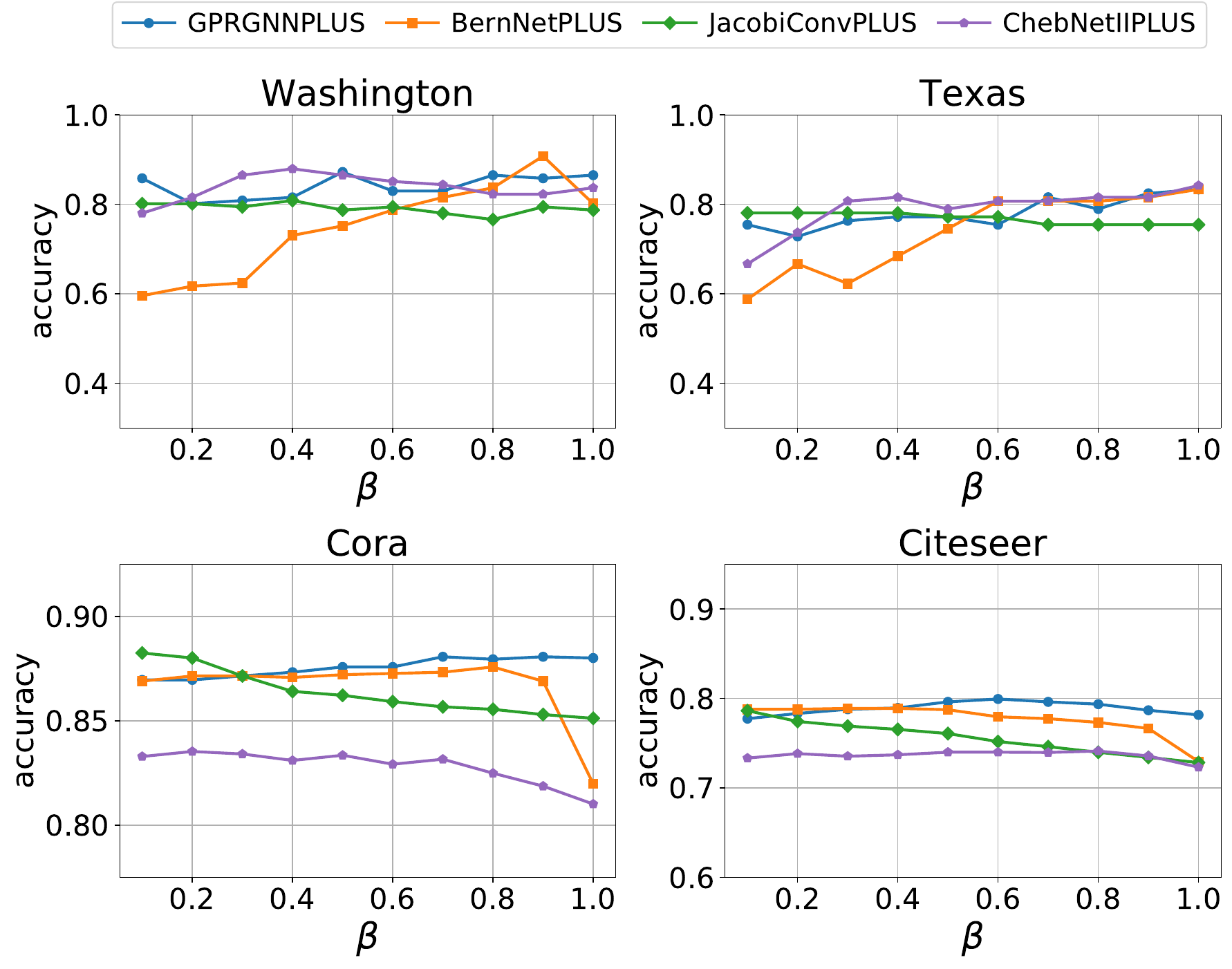}
	\caption{Effect of hyperparameter $\beta$ on model performance.}
    \label{fig:beta}
  \label{fig:hyperparameter}
\end{figure}

\subsection{Efficiency Studies (\textbf{RQ3})}
We assess the efficiency of our method in terms of both monetary and time costs. Leveraging low-cost GPT-4o-mini API calls for homophily estimation, the financial overhead remains negligible—under \$0.2 per dataset as shown in Table~\ref{tab:cost}. For time efficiency, Table~\ref{tab:time} compares our method with baseline spectral GNNs in per-epoch and total training time, showing no significant runtime increase. The only extra cost stems from constructing heterophily-aware basis functions, which are precomputed once per polynomial order and reused across layers. During training, these basis vectors are simply loaded, adding minimal overhead. Notably, our method requires no fine-tuning or local model deployment, making it lightweight and practical.

\subsection{Parameter Sensitivity Analysis (\textbf{RQ4})}
Our method introduces a key hyperparameter $\beta$ to balance the original polynomial basis and the heterophily-aware basis informed by predicted homophily. Smaller $\beta$ values emphasize the heterophily-aware component, while larger values favor the original basis. Figure~\ref{fig:beta} shows the performance sensitivity to $\beta$ across four representative datasets. The results reveal diverse trends. For example, in Washington and Texas, BernNetPLUS shows an upward trend, indicating its strong filter capacity benefits from the original basis. In contrast, the other three methods maintain stable performance in Washington and Texas, suggesting robustness to $\beta$ and complementary contributions from both bases. On datasets like Cora and Citeseer, all models perform better with smaller $\beta$, implying that incorporating heterophily-aware basis functions—guided by LLM-estimated homophily—enhances spectral expressiveness even in largely homophilic graphs.

%% file: 6_conclusion.tex
\section{Conclusion}

This paper is the first to investigate the potential of LLMs in enhancing spectral GNNs. Specifically, we propose a novel method that utilizes LLM-predicted homophily ratio to construct a heterophily basis, which is then integrated into existing spectral GNN architectures, enabling spectral GNNs to handle both homophilic and heterophilic graphs better. Extensive experiments on multiple benchmark datasets demonstrate that our method consistently improves the performance of various spectral GNNs, while incurring minimal computational and monetary overhead. Besides, these results highlight the promising synergy between LLMs and spectral graph learning, and open up new directions for future research at the intersection of larger language models and graph neural networks.


%% file: 7_appendix.tex
\section{More Experimental Details}

\subsection{Datasets}
\begin{itemize}
\item \textbf{Cora, Citeseer, and Pubmed} \cite{graphedit} are three widely used citation network datasets in graph learning. In each dataset, nodes represent papers and edges indicate citation links between them. Node features are bag-of-words representations of the papers, and labels correspond to document categories. 

\item  \textbf{Cornell, Texas, and Wisconsin} \cite{CS-TAG} are webpage networks collected from computer science departments of different universities. In these datasets, nodes represent web pages and edges denote hyperlinks between them. We extract node features by encoding the raw webpage content using the pre-trained \texttt{bert-base-uncased} model.

\item  \textbf{Children and History} \cite{CS-TAG} are two subsets of the Amazon-Books dataset, containing books labeled with the second-level categories “Children” and “History”, respectively. Nodes represent books, and edges indicate frequent co-purchase or co-view relationships. We use the book titles and descriptions as textual attributes, encoded with the pre-trained \texttt{roberta-base} model. 

\item  \textbf{Fitness} \cite{CS-TAG} is a subset of the Amazon-Sports dataset, containing items labeled with the second-level category “Fitness”. Nodes represent fitness-related products, and edges indicate frequent co-purchase or co-view relationships. We use the product titles and descriptions as textual features, encoded with the pre-trained \texttt{roberta-base} model. 

\end{itemize}

\subsection{Baselines}
To facilitate understanding of the models used in our experiments, we briefly introduce the backbone polynomial spectral GNN and the baseline GNN methods evaluated in this paper:

\begin{itemize}
    \item \textbf{GCN}~\cite{GCN} introduces a spectral-based convolution operation on graphs, aggregating information from immediate neighbors with a normalized adjacency matrix.

    \item \textbf{GraphSAGE}~\cite{GraphSAGE} samples and aggregates features from a node's local neighborhood using functions like mean or LSTM, enabling inductive learning on large-scale graphs.
    
    \item \textbf{GAT}~\cite{GAT} leverages self-attention mechanisms to assign different importance weights to neighbors, enhancing the flexibility of node representation learning.
    
    \item \textbf{APPNP}~\cite{PPNP} decouples feature transformation and propagation by applying personalized PageRank-based diffusion on top of a neural predictor.

    \item  \textbf{GPRGNN} \cite{GPR-GNN} extends APPNP by directly assigning a learnable coefficient to each order of the normalized adjacency matrix.

    \item \textbf{BernNet} \cite{BernNet} expresses the filtering operation with Bernstein polynomials and forces all coefficients to be positive.

    \item \textbf{JacobiConv} \cite{JacobiConv} proposes a Jacobi basis to adapt a wide range of weight functions due to its orthogonality and flexibility.

    \item \textbf{ChebNetII} \cite{ChebNetII} addresses overfitting and the Runge phenomenon by enhancing the original Chebyshev polynomial approximation, reparameterizing the filter response as a learnable coefficient.

    \item \textbf{TFE-GNN}~\cite{TFEGNN} introduces a triple filter ensemble to combine low-pass, high-pass, and initial features, enabling adaptive extraction of homophily and heterophily. 

    \item \textbf{UniFilter}~\cite{UniFilter} unifies heterophilic and homophilic bases to construct a universal polynomial filter that alleviates over-smoothing and over-squashing.
\end{itemize}

\subsection{Hyper-parameter Settings}
In all experiments, the number of training epochs is set to 1000, and the learning rate is fixed at 0.0001. A three-layer MLP with a hidden dimension of 256 is used for feature transformation. For each dataset, 100 edges are sampled. The key hyperparameter $\beta$ is tuned over the range $\{0.0, 0.1, 0.2, \dots, 1.0\}$. The optimal $\beta$ values for each method on each dataset are shown in Table 1. 
\begin{table*}[htbp]
\centering
\caption{Optimal $\beta$ values for each method on each dataset.}
\begin{tabular}{lccccccccc}
\toprule
Dataset & Washington & Cornell & Texas & Children & History & Fitness & Cora & Citeseer & Pubmed \\
\midrule
GPRGNNPLUS     & 0.5 & 1.0 & 1.0 & 1.0 & 1.0 & 0.1 & 0.7 & 0.6 & 0.3 \\
BernNetPLUS    & 0.9 & 0.5 & 1.0 & 0.2 & 0.1 & 0.2 & 0.8 & 0.3 & 0.5 \\
JacobiConvPLUS & 0.4 & 0.0 & 0.0 & 0.8 & 0.7 & 0.6 & 0.1 & 0.1 & 0.1 \\
ChebnetIIPLUS  & 0.4 & 0.8 & 1.0 & 0.1 & 0.1 & 0.0 & 0.2 & 0.8 & 1.0 \\
\bottomrule
\end{tabular}
\end{table*}